\documentclass[11pt,a4paper]{article}
\usepackage[hyperref]{emnlp-ijcnlp-2019}
\usepackage{times}
\usepackage{latexsym}

\usepackage{url}

\usepackage{graphicx}
\usepackage{amsmath}
\usepackage{color}
\aclfinalcopy 

\title{Transfer in Deep Reinforcement Learning using Knowledge Graphs}

\author{Prithviraj Ammanabrolu \\
  School of Interactive Computing \\
  Georgia Institute of Technology \\
  Atlanta, GA \\
  {\tt raj.ammanabrolu@gatech.edu} \\\And
  Mark O. Riedl \\
  School of Interactive Computing \\
  Georgia Institute of Technology \\
  Atlanta, GA \\
  {\tt riedl@cc.gatech.edu}}

\date{}

\begin{document}
\maketitle
\begin{abstract}
Text adventure games, in which players must make sense of the world through text descriptions and declare actions through text descriptions, 
provide a stepping stone toward grounding action in language.
Prior work has demonstrated that using a knowledge graph as a state representation and question-answering to pre-train a deep Q-network facilitates faster control policy learning.
In this paper, we explore the use of knowledge graphs as a representation for domain knowledge transfer for training text-adventure playing reinforcement learning agents.
Our methods are tested across multiple computer generated and human authored games, varying in domain and complexity, and demonstrate that our transfer learning methods let us learn a higher-quality control policy faster.
\end{abstract}

\section{Introduction}
Text adventure games, in which players must make sense of the world through text descriptions and declare actions through natural language, can provide a stepping stone toward more real-world environments where agents must communicate to understand the state of the world and affect change in the world.
Despite the steadily increasing body of research on text-adventure games \citep{bordes,He2015,Narasimhan2015,fulda-ijcai2017,Haroush2018,Cote2018,tao18,ammanabrolu}, and in addition to the ubiquity of deep reinforcement learning applications \citep{Parisotto2016ActorMimicDM,zambaldi2018deep}, 
teaching an agent to play text-adventure games remains a challenging task.
Learning a control policy for a text-adventure game requires a significant amount of exploration, resulting in training runs that take hundreds of thousands of simulations \citep{Narasimhan2015,ammanabrolu}.

One reason that text-adventure games require so much exploration is that most deep reinforcement learning algorithms are trained on a task without a real prior.
In essence, the agent must learn everything about the game from only its interactions with the environment.
Yet, text-adventure games make ample use of commonsense knowledge (e.g., an axe can be used to cut wood) and genre themes (e.g., in a horror or fantasy game, a coffin is likely to contain a vampire or other undead monster).
This is in addition to the challenges innate to the text-adventure game itself---games are puzzles---which results in inefficient training.

\citet{ammanabrolu} developed a reinforcement learning agent that modeled the text environment as a knowledge graph and achieved state-of-the-art results on simple text-adventure games provided by the TextWorld~\cite{Cote2018} environment.
They observed that a simple form of transfer from very similar games greatly improved policy training time.
However, 
games beyond the toy TextWorld environments are
beyond the reach of state-of-the-art techniques.

In this paper, we explore the use of  knowledge graphs and associated neural embeddings as a medium for domain transfer to improve training effectiveness on new text-adventure games.
Specifically, we explore transfer learning at multiple levels and across different dimensions.
We first look at the effects of playing a text-adventure game given a strong prior in the form of a knowledge graph extracted from generalized textual walk-throughs of interactive fiction as well as those made specifically for a given game.
Next, we explore the transfer of control policies in deep Q-learning (DQN) by pre-training portions of a deep Q-network using question-answering and by DQN-to-DQN parameter transfer between games.
We evaluate these techniques on two different sets of human authored and computer generated games, demonstrating that our transfer learning methods enable us to learn a higher-quality control policy faster.


\section{Background and Related Work}

Text-adventure games, in which an agent must interact with the world entirely through natural language, provide us with two challenges that have proven difficult for deep reinforcement learning to solve 
\citep{Narasimhan2015,Haroush2018,ammanabrolu}:
(1)~The agent must act based only on potentially incomplete textual descriptions of the world around it.
The world is thus partially observable, as the agent does not have access to the state of the world at any stage.
(2)~the action space is combinatorially large---a consequence of the agent having to declare commands in natural language.
These two problems together have kept commercial text adventure games out of the reach of existing deep reinforcement learning methods, especially given the fact that most of these methods attempt to train on a particular game from scratch.

Text-adventure games can be treated as partially observable Markov decision processes (POMDPs).
This can be represented as a 7-tuple of $\langle S,T,A, \Omega , O,R, \gamma\rangle$: the set of environment states, conditional transition probabilities between states, words used to compose text commands, observations, conditional observation probabilities, the reward function, and the discount factor respectively~\cite{Cote2018}.

Multiple recent works have explored the challenges associated with these games \citep{bordes,He2015,Narasimhan2015,fulda-ijcai2017,Haroush2018,Cote2018,tao18,ammanabrolu}. 
\citet{Narasimhan2015} introduce the LSTM-DQN, which learns to score the action verbs and corresponding objects separately and then combine them into a single action.
\citet{He2015} propose the Deep Reinforcement Relevance Network that consists of separate networks to encode state and action information, with a final Q-value for a state-action pair that is computed between a pairwise interaction function between these.
\citet{Haroush2018} present the Action Elimination Network (AEN), which restricts actions in a state to the top-k most likely ones, using the emulator's feedback.
\citet{hausknecht} design an agent that uses multiple modules to identify a general set of game play rules for text games across various domains.
None of these works study how to transfer policies between different text-adventure games in any depth and so there exists a gap between the two bodies of work.

Transferring policies across different text-adventure games requires implicitly learning a mapping between the games' state and action spaces.
The more different the domain of the two games, the harder this task becomes.
Previous work~\cite{ammanabrolu} introduced the use of knowledge graphs and question-answering pre-training to aid in the problems of partial observability and a combinatorial action space.
This work made use of a system called TextWorld~\cite{Cote2018} that uses grammars to generate a series of similar (but not exact same) games. 
An oracle was used to play perfect games and the traces were used to pre-train portions of the agent's network responsible for encoding the observations, graph, and actions.
Their results show that this form of pre-training improves the quality of the policy at convergence it does not show a significant improvement in the training time required to reach convergence.
Further, it is generally unrealistic to have a corpus of very similar games to draw from.
We build on this work, and explore modifications of this algorithm that would enable more efficient transfer in text-adventure games.




Work in transfer in reinforcement learning has explored the idea of transferring skills \citep{Konidaris2007BuildingPO,konidaris12} or transferring value functions/policies \citep{Liu2006ValueFunctionBasedTF}.
Other approaches attempt transfer in model-based reinforcement learning \citep{Taylor2008TransferringIF,Nguyen2012TransferringEI,Gasic2013POMDPbasedDM,Wang2015LearningDD,joshi}, though traditional approaches here rely heavily on hand crafting state-action mappings across domains.
\citet{narasimhanGround} learn to play games by predicting mappings across domains using a both deep Q-networks and value iteration networks, finding that that grounding the game state using natural language descriptions of the game itself aids significantly in transferring useful knowledge between domains.

In transfer for deep reinforcement learning, \citet{Parisotto2016ActorMimicDM} propose the Actor-Mimic network which learns from expert policies for a source task using policy distillation and then initializes the network for a target task using these parameters.
\citet{yinpan} also use policy distillation, using task specific features as inputs to a multi-task policy network and use a hierarchical experience sampling method to train this multi-task network.
Similarly, \citet{Rusu2016ProgressiveNN} attempt to transfer parameters by using frozen parameters trained on source tasks to help learn a new set of parameters on target tasks.
\citet{Rajendran2017AttendAA} attempt something similar but use attention networks to transfer expert policies between tasks.
These works, however, do not study the requirements for enabling efficient transfer for tasks rooted in natural language, nor do they explore the use of knowledge graphs as a state representation.


\section{Knowledge Graphs for DQNs}
\label{sec:kg}

A knowledge graph is a directed graph formed by a set of semantic, or RDF, triples in the form of $\langle subject, relation, object\rangle$---for example, $\langle vampires, are, undead\rangle$.
We follow the open-world assumption that what is not in our knowledge graph can either be true or false.

\citet{ammanabrolu} introduced the Knowledge Graph DQN (KG-DQN) and touched on some aspects of transfer learning, showing that pre-training portions of the deep Q-network using question answering system on perfect playthroughs of a game increases the quality of the learned control policy for a generated text-adventure game.
We build on this work and use KG-DQN to explore transfer with both knowledge graphs and network parameters.
Specifically we seek to transfer skills and knowledge from (a)~static text documents describing game play and (b)~from playing one text-adventure game to a second complete game in in the same genre (e.g., horror games).
The rest of this section describes KG-DQN in detail and summarizes our modifications.\footnote{We use the implementation of KG-DQN found at \url{https://github.com/rajammanabrolu/KG-DQN}}

For each step that the agent takes, it automatically extracts a set of RDF triples from the received observation through the use of OpenIE \citep{Angeli2015} in addition to a few rules to account for the regularities of text-adventure games.
The graph itself is more or less a map of the world, with information about objects' affordances and attributes linked to the rooms that they are place in in a map.
The graph also makes a distinction with respect to items that are in the agent's possession or in their immediate surrounding environment.
We make minor modifications to the rules used in \citet{ammanabrolu} to better achieve such a graph in general interactive fiction environments.

The agent also has access to all actions accepted by the game's parser, following \citet{Narasimhan2015}.
For general interactive fiction environments, we develop our own method to extract this information.
This is done by extracting a set of templates accepted by the parser, with the objects or noun phrases in the actions replaces with a OBJ tag.
An example of such a template is "place OBJ in OBJ".
These OBJ tags are then filled in by looking at all possible objects in the given vocabulary for the game.
This action space is of the order of $A=\mathcal{O}(|V| \times |O|^2)$ where $V$ is the number of action verbs, and $O$ is the number of distinct objects in the world that the agent can interact with.
As this is too large a space for a RL agent to effectively explore, the knowledge graph is used to prune this space by ranking actions based on their presence in the current knowledge graph and the relations between the objects in the graph as in~\cite{ammanabrolu}

\begin{figure}[tb]
    \centering
    \includegraphics[width=0.8\linewidth]{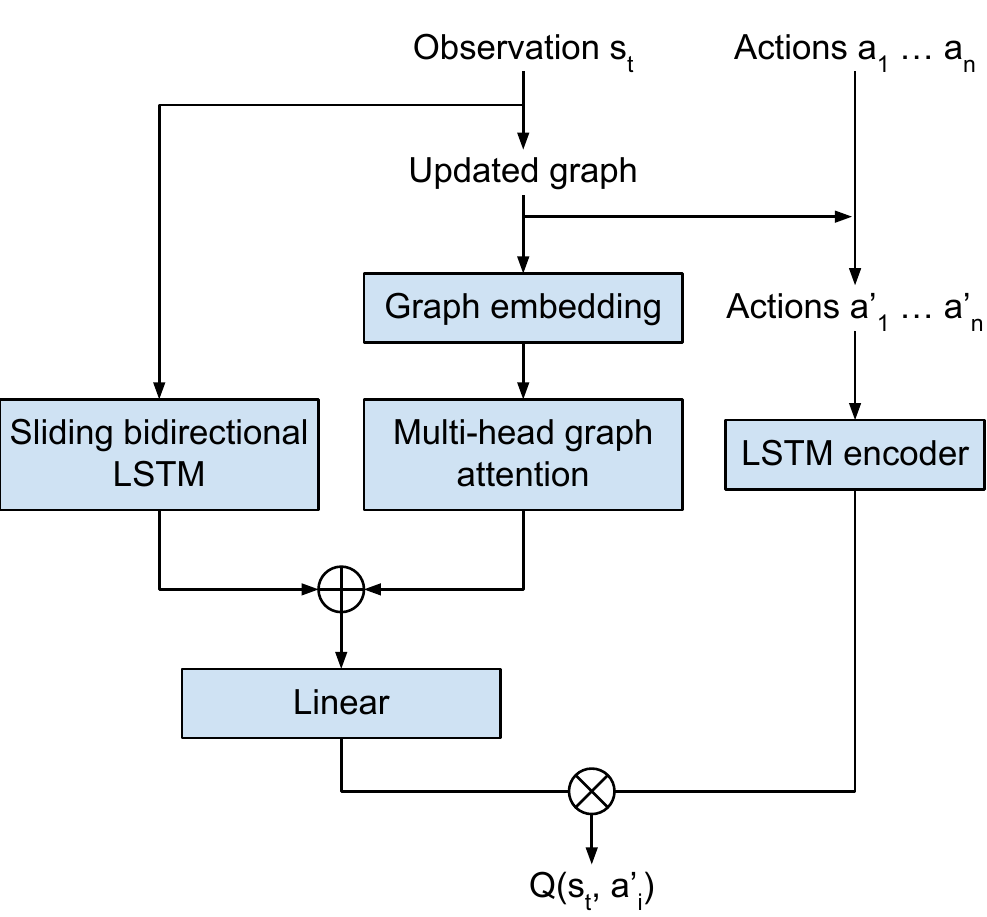}
    \caption{The KG-DQN architecture.}
    \label{fig:kg-dqn}
\end{figure}

The architecture for the deep Q-network consists of two separate neural networks---encoding state and action separately---with the final $Q$-value for a state-action pair being the result of a pair-wise interaction function between the two (Figure~\ref{fig:kg-dqn}).
We train with a standard DQN training loop;
the policy is determined by the $Q$-value of a particular state-action pair, which is updated using the Bellman equation \citep{Sutton2018}:
\begin{equation}
\begin{aligned}
    Q_{t+1}(s_{t+1}, &a_{t+1}) =\\
   & E[r_{t+1} + \gamma \max_{a \in A_{t}} Q_t(s, a)|s_t,a_t]
\end{aligned}
\end{equation}
where $\gamma$ refers to the discount factor and $r_{t+1}$ is the observed reward. 
The whole system is trained using prioritized experience replay \citet{Lin1993}, a modified version of $\epsilon$-greedy learning, and a temporal difference loss that is computed as:
\begin{equation}
\begin{aligned}
    L(\theta) =& r_{k+1} +\\ 
    &\gamma \max_{\mathbf{a} \in \mathbf{A_{k+1}}}Q(\mathbf{s_t}, \mathbf{a}; \theta) - Q(\mathbf{s_t}, \mathbf{a_t}; \theta)
\end{aligned}
\end{equation}
where $\mathbf{A_{k+1}}$ represents the action set at step $k$ + 1 and $\mathbf{s_t,a_t}$ refer to the encoded state and action representations respectively.



\section{Knowledge Graph Seeding}
\label{sec:seeding}

In this section we consider the problem of 
transferring a knowledge graph from a static text resource to a DQN---which we refer to as {\em seeding}.
%
KG-DQN uses a knowledge graph as a state representation and also to prune the action space.
This graph is built up over time, through the course of the agent's exploration.
When the agent first starts the game, however, this graph is empty and does not help much in the action pruning process.
The agent thus wastes a large number of steps near the beginning of each game exploring ineffectively.

The intuition behind seeding the knowledge graph from another source is to give the agent a prior on which actions have a higher utility and thereby enabling more effective exploration.
Text-adventure games typically belong to a particular genre of storytelling---e.g., horror, sci-fi, or soap opera---and an agent is at a distinct disadvantage if it doesn't have any genre knowledge.
Thus, the goal of seeding is to give the agent a strong prior.

\begin{figure}
    \centering
    \includegraphics[width=0.9\linewidth]{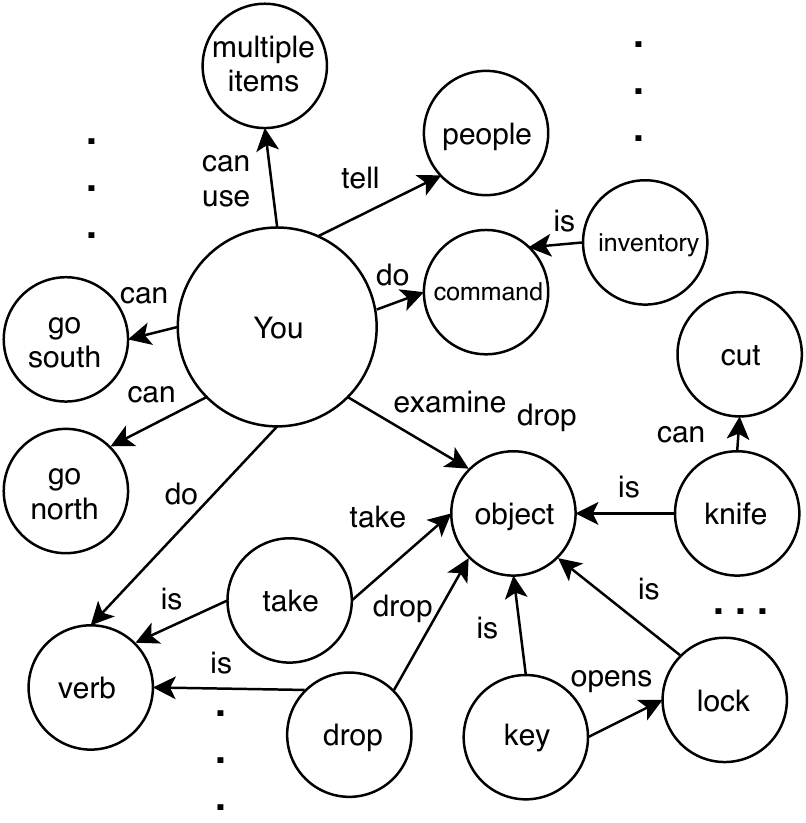}
    \caption{Select partial example of what a seed knowledge graph looks like. Ellipses indicate other similar entities and relations not shown.}
    \label{fig:seed_kg}
\end{figure}

This seed knowledge graph is extracted from online general text-adventure guides as well as game/genre specific guides when available.\footnote{An example of a guide we use is found here \url{http://www.microheaven.com/IFGuide/step3.html}}
The graph is extracted from this the guide using a subset of the rules described in Section~\ref{sec:kg} used to extract information from the game observations, with the remainder of the RDF triples coming from OpenIE.
There is no map of rooms in the environment that can be built, but it is possible to extract information regarding affordances of frequently occurring objects as well as common actions that can be performed across a wide range of text-adventure games.
This extracted graph is thus potentially disjoint, containing only this generalizable information, in contrast to the graph extracted during the rest of the exploration process.
An example of a graph used to seed KG-DQN is given in Fig.~\ref{fig:seed_kg}.
The KG-DQN is initialized with this knowledge graph.


\section{Task Specific Transfer}


The overarching goal of transfer learning in text-adventure games is to be able to train an agent on one game and use this training on to improve the learning capabilities of another.
There is growing body of work on improving training times on target tasks by transferring network parameters trained on source tasks \citep{Rusu2016ProgressiveNN, yinpan, Rajendran2017AttendAA}.
Of particular note is the work by \citet{Rusu2016ProgressiveNN}, where they train a policy on a source task and then use this to help learn a new set of parameters on a target task.
In this approach, decisions made during the training of the target task are jointly made using the frozen parameters of the transferred policy network as well as the current policy network.

Our system first trains a question-answering system \citep{chen2017reading} using traces given by an oracle, as in Section~\ref{sec:seeding}.
For commercial text-adventure games, these traces take the form of state-action pairs generated using perfect walkthrough descriptions of the game found online as described in Section~\ref{sec:seeding}.

We use the parameters of the question-answering system to pre-train portions of the deep Q-network for a different game within in the same domain.
The portions that are pre-trained are the same parts of the architecture as in \citet{ammanabrolu}.
This game is referred to as the {\bf source task}.
The seeding of the knowledge graph is not strictly necessary but given that state-of-the-art DRL agents cannot complete real games, this makes the agent more effective at the source task.

We then transfer the knowledge and skills acquired from playing the source task to another game from the same genre---the {\bf target task}.
The parameters of the deep Q-network trained on the source game are used to initialize a new deep Q-network for the target task.
All the weights indicated in the architecture of KG-DQN as shown in Fig.~\ref{fig:kg-dqn} are transferred.
Unlike \citet{Rusu2016ProgressiveNN}, we do not freeze the parameters of the deep Q-network trained on the source task nor use the two networks to jointly make decisions but instead just use it to initialize the parameters of the target task deep Q-network.
This is done to account for the fact that although graph embeddings can be transferred between games, the actual graph extracted from a game is non-transferable due to differences in structure between the games.


\section{Experiments}

We test our system on two separate sets of games in different domains using the Jericho and TextWorld frameworks~\citep{jericho, Cote2018}.
The first set of games is ``slice of life'' themed and contains games that involve mundane tasks usually set in textual descriptions of normal houses.
The second set of games is ``horror'' themed and contains noticeably more difficult games with a relatively larger vocabulary size and action set, non-standard fantasy names, etc.
We choose these domains because of the availability of games in popular online gaming communities, the degree of vocabulary overlap within each theme, and overall structure of games in each theme.
Specifically, there must be at least three games in each domain: at least one game to train the question-answering system on, and two more to train the parameters of the source and target task deep Q-networks.
A summary of the statistics for the games is given in Table~\ref{table:games}.
Vocabulary overlap is calculated by measuring the percentage of overlap between a game's vocabulary and the domain's vocabulary, i.e. the union of the vocabularies for all the games we use within the domain.
We observe that in both of these domains, the complexity of the game increases steadily from the game used for the question-answering system to the target and then source task games.
\begin{table*}[]
\footnotesize
\centering
\caption{Game statistics}
\label{table:games}
\begin{tabular}{l|cc|ccc}
\hline
\multicolumn{1}{c|}{}       & \multicolumn{2}{c|}{Slice of life}    & \multicolumn{3}{c}{Horror}                                  \\ \hline
\multicolumn{1}{c|}{}       & QA/Source & \multicolumn{1}{c|}{Target} & QA & Source & \multicolumn{1}{c}{Target} \\ 
\multicolumn{1}{c|}{}       & {\em TextWorld} & \multicolumn{1}{c|}{\em 9:05} & {\em Lurking Horror} & {\em Afflicted} & \multicolumn{1}{l}{\em Anchorhead} \\ \hline
Vocab size                  &    788       &          297               &      773       &     761&  2256                               \\
Branching factor            &     122      &         677                  &       -        &     947 & 1918                                \\
Number of rooms             &     10      &          7              &        25      &      18     & 28                             \\
Completion steps       &      5     &           25            &       289      &     20&   39                              \\
Words per obs.           &     65.1      &            45.2               &        68.1     &   81.2      &               114.2                  \\
New triples per obs. &     6.4      &              4.1             &        -        &    12.6   & 17.0 \\
\% Vocab overlap &   19.70       &         21.45             &           22.80          &   14.40    &    66.34 \\
Max. aug. reward & 5 & 27 & - & 21 & 43
\end{tabular}
\end{table*}

We perform ablation tests within each domain, mainly testing the effects of transfer from seeding, oracle-based question-answering, and source-to-target parameter transfer.
Additionally, there are a couple of extra dimensions of ablations that we study, specific to each of the domains and explained below.
All experiments are run three times using different random seeds.
For all the experiments we report metrics known to be important for transfer learning tasks \citep{Taylor2009TransferLF,narasimhanGround}: average reward collected in the first 50 episodes (init. reward), average reward collected for 50 episodes after convergence (final reward), and number of steps taken to finish the game for 50 episodes after convergence (steps).
For the metrics tested after convergence, we set $\epsilon=0.1$ following both \citet{Narasimhan2015} and \citet{ammanabrolu}.
We use similar hyperparameters to those reported in \cite{ammanabrolu} for training the KG-DQN with action pruning, with the main difference being that we use 100 dimensional word embeddings instead of 50 dimensions for the horror genre.

\subsection{Slice of Life Experiments}

\begin{figure}[tb]
    \centering
    \includegraphics[width=0.75\linewidth]{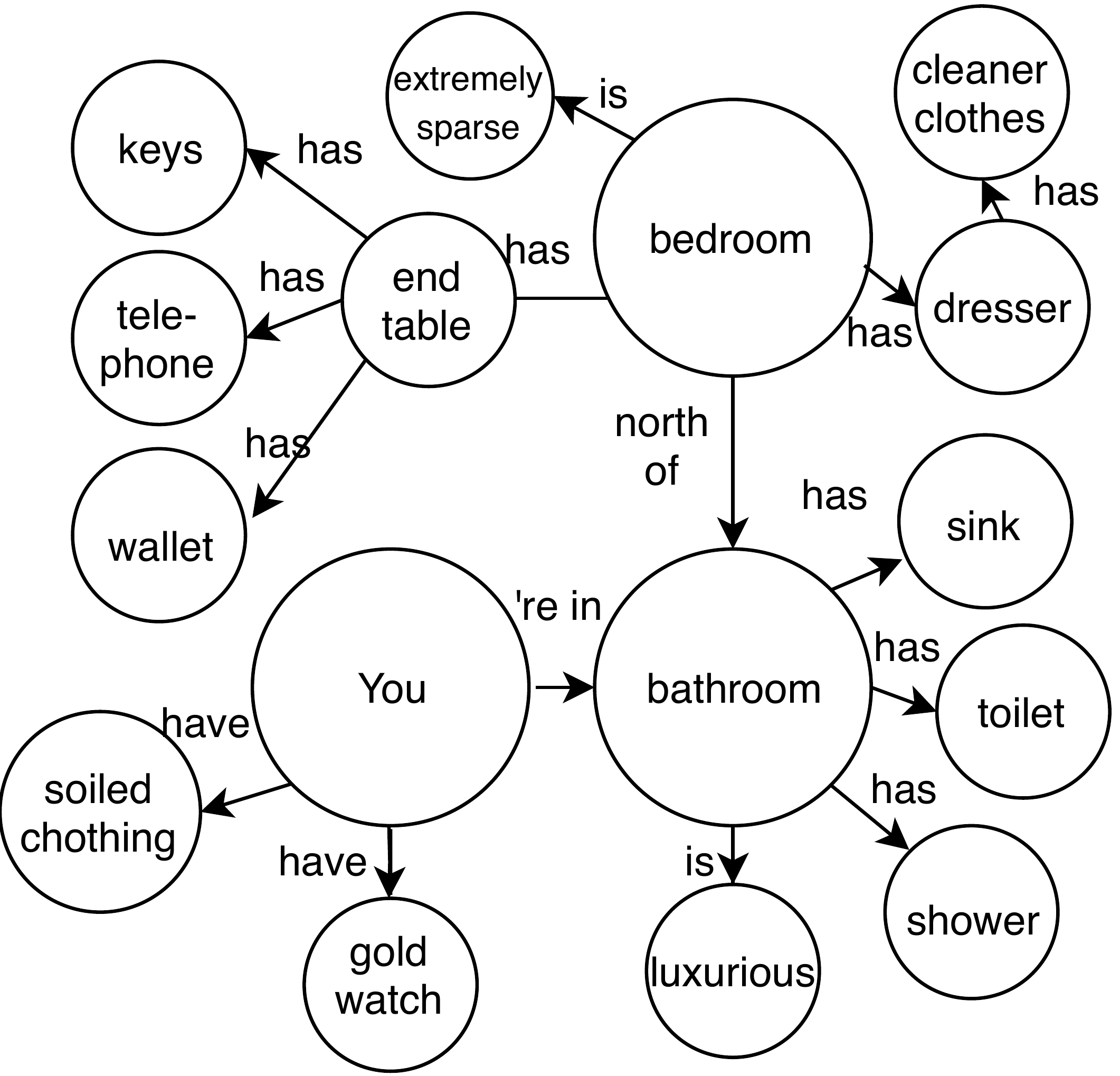}
    \includegraphics[width=\linewidth]{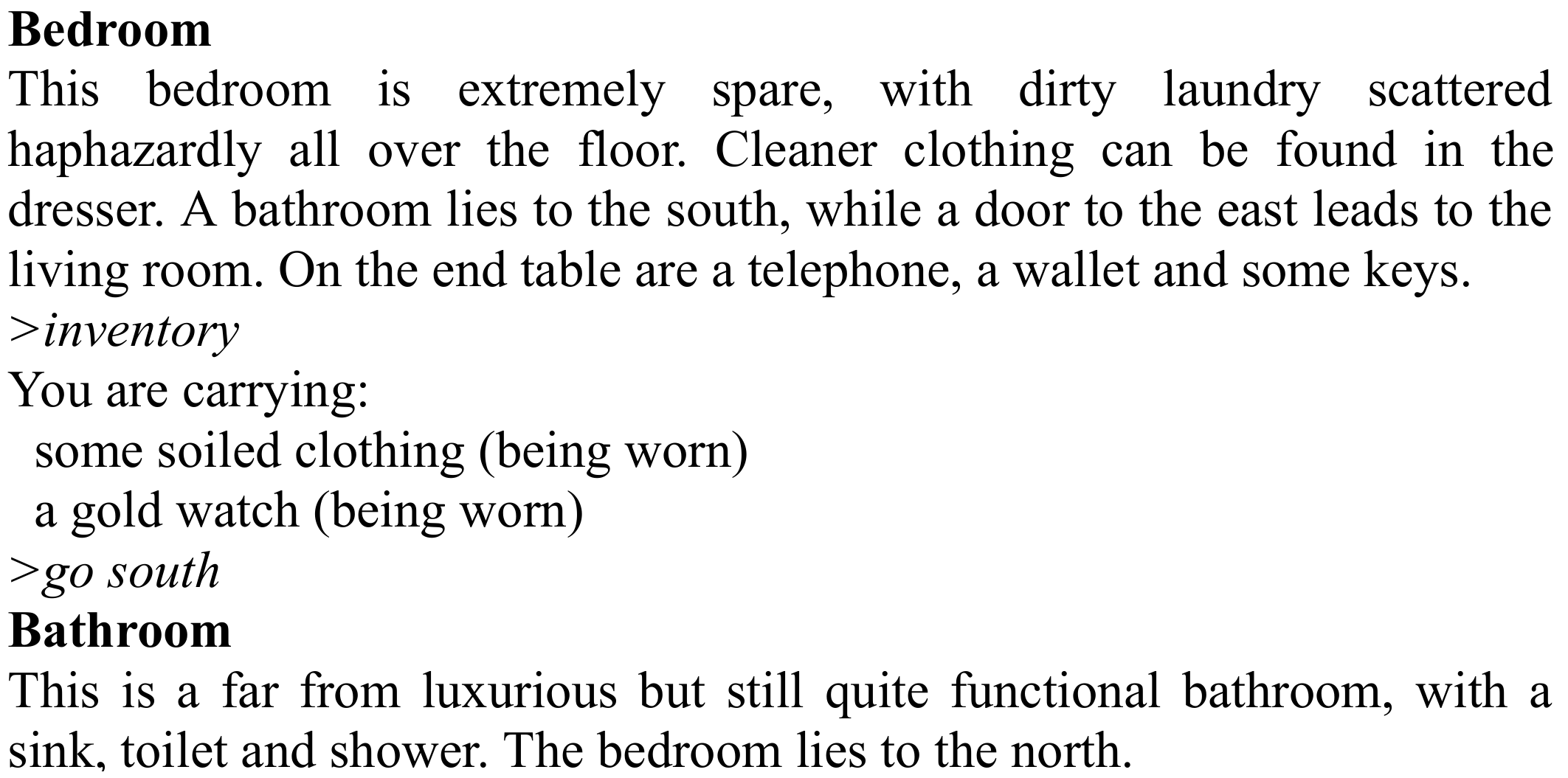}
    \caption{Partial unseeded knowledge graph example given observations and actions in the game {\em 9:05}.}
    \label{fig:kg-sol}
\end{figure}
\begin{figure}[tb]
    \centering
    \includegraphics[width=0.75\linewidth]{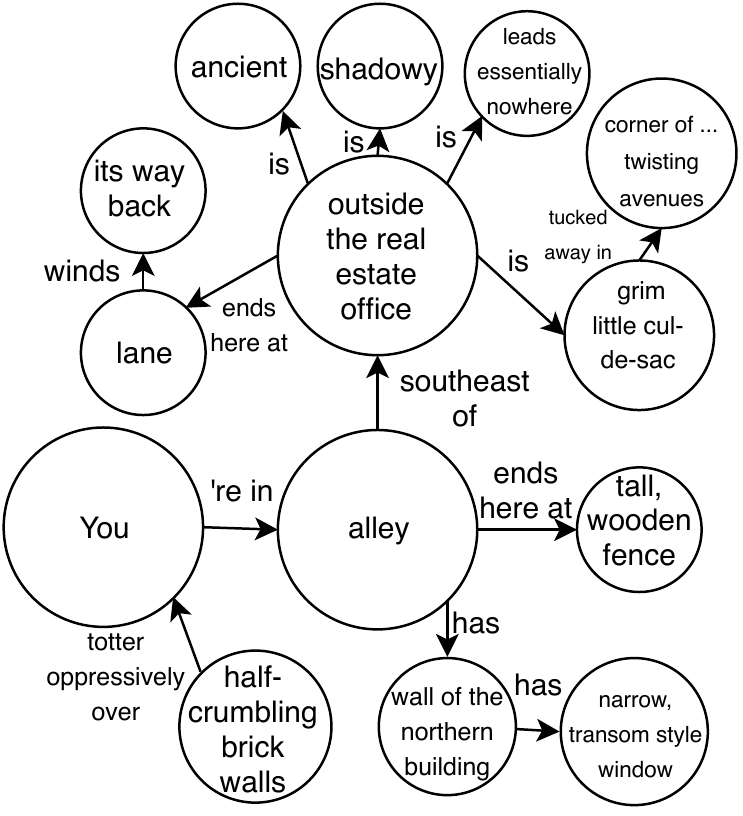}
    \includegraphics[width=\linewidth]{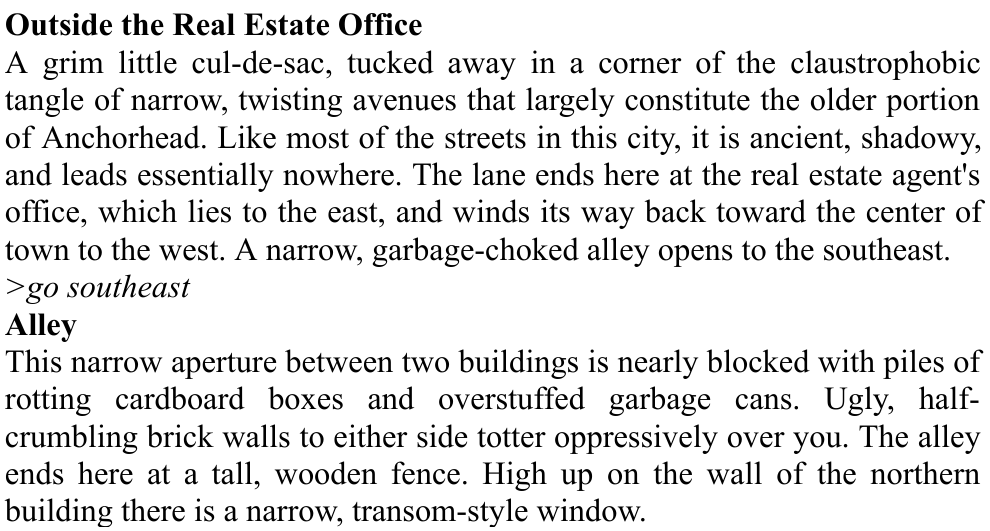}
    \caption{Partial unseeded knowledge graph example given observations and actions in the game {\em Anchorhead}.}
    \label{fig:kg-horror}
\end{figure}

TextWorld uses a grammar to generate similar games.
Following \citet{ammanabrolu}, we use TextWorld's ``home'' theme to generate the games for the question-answering system.
TextWorld is a framework that uses a grammar to randomly generate game worlds and quests.
This framework also gives us information such as instructions on how to finish the quest, and a list of actions that can be performed at each step based on the current world state.
We do not let our agent access this additional solution information or admissible actions list.
Given the relatively small quest length for TextWorld games---games can be completed in as little as 5 steps---we generate 50 such games and partition them into train and test sets in a 4:1 ratio.
The traces are generated on the training set, and the question-answering system is evaluated on the test set.

We then pick a random game from the test set to train our source task deep Q-network for this domain.
For this training, we use the reward function provided by TextWorld: +1 for each action taken that moves the agent closer to finishing the quest; -1 for each action taken that extends the minimum number of steps needed to finish the quest from the current stage; 0 for all other situations.

We choose the game, {\em 9:05}\footnote{\footnotesize{\url{ https://ifdb.tads.org/viewgame?id=qzftg3j8nh5f34i2}}} as our target task game due to similarities in structure in addition to the vocabulary overlap.
Note that there are multiple possible endings to this game and we pick the simplest one for the purpose of training our agent.

\subsection{Horror Experiments}

For the horror domain, we choose {\em Lurking Horror}\footnote{\url{https://ifdb.tads.org/viewgame?id=jhbd0kja1t57uop}} to train the question-answering system on.
The source and target task games are chosen as {\em Afflicted}\footnote{\url{https://ifdb.tads.org/viewgame?id=epl4q2933rczoo9x}} and {\em Anchorhead}\footnote{\url{https://ifdb.tads.org/viewgame?id=op0uw1gn1tjqmjt7}} respectively.
However, due to the size and complexity of these two games 
some modifications to the games are required for the agent to be able to effectively solve them.
We partition each of these games and make them smaller by reducing the final goal of the game to an intermediate checkpoint leading to it.
This checkpoints were identified manually using walkthroughs of the game; each game has a natural intermediate goal.
For example, {\em Anchorhead} is segmented into 3 chapters in the form of objectives spread across 3 days, of which we use only the first chapter.
The exact details of the games after partitioning is described in Table~\ref{table:games}.
For Lurking Horror, we report numbers relevant for the oracle walkthrough.
We then pre-prune the action space and use only the actions that are relevant for the sections of the game that we have partitioned out.
The majority of the environment is still available for the agent to explore but the game ends upon completion of the chosen intermediate checkpoint.

\subsection{Reward Augmentation}
\label{sec:reward-aug}

\begin{figure}
    \centering
    \includegraphics[width=\linewidth]{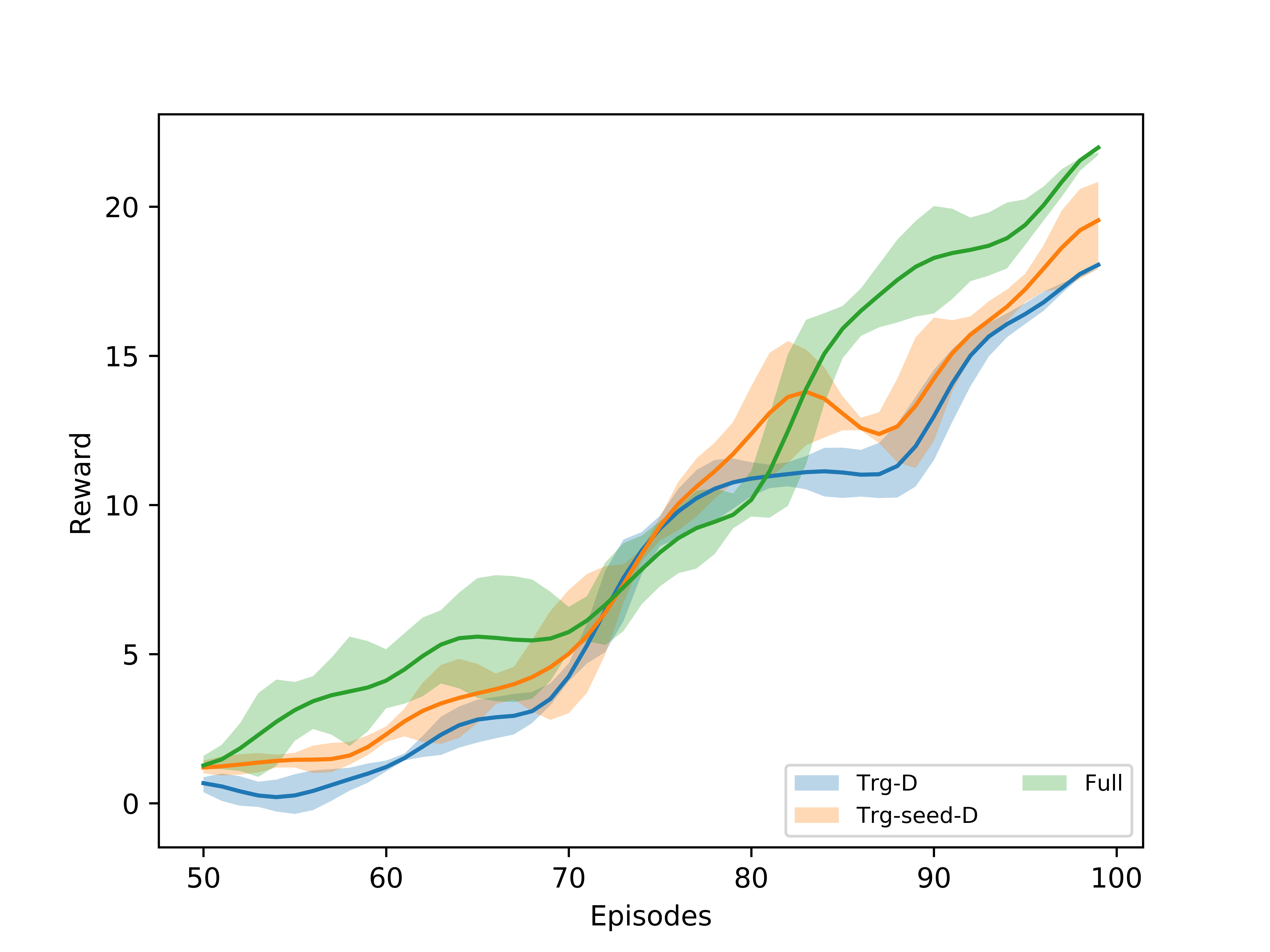}
    \caption{Reward curve for select experiments in the slice of life domain.}
    \label{fig:reward-sol}
\end{figure}
The combined state-action space for a commercial text-adventure game is quite large and the corresponding reward function is very sparse in comparison.
The default, implied reward signal is to receive positive value upon completion of the game, and no reward value elsewhere.
This is problematic from an experimentation perspective as text-adventure games are too complex for even state-of-the-art deep reinforcement learning agents to complete.
Even using transfer learning methods, a sparse reward signal usually results in ineffective exploration by the agent.

To make experimentation feasible, we augment the reward to give the agent a dense reward signal. 
Specifically, we use an oracle to generate state-action traces (identical to how as when training the question-answering system).
An oracle is an agent that is capable of playing and finishing a game perfectly in the least number of steps possible.
The state-action pairs generated using perfect walkthroughs of the game are then used as checkpoints and used to give the agent additional reward.
If the agent encounters any of these state-action pairs when training, i.e. performs the right action given a corresponding state, it receives a proportional reward in addition to the standard reward built into the game.
This reward is scaled based on the game and is designed to be less than the smallest reward given by the original reward function to prevent it from overpowering the built-in reward.
We refer to agents using this technique as having ``dense'' reward and ``sparse'' reward otherwise.
The agent otherwise receives no information from the oracle about how to win the game.

\section{Results/Discussion}

\begin{figure}
    \centering
    \includegraphics[width=\linewidth]{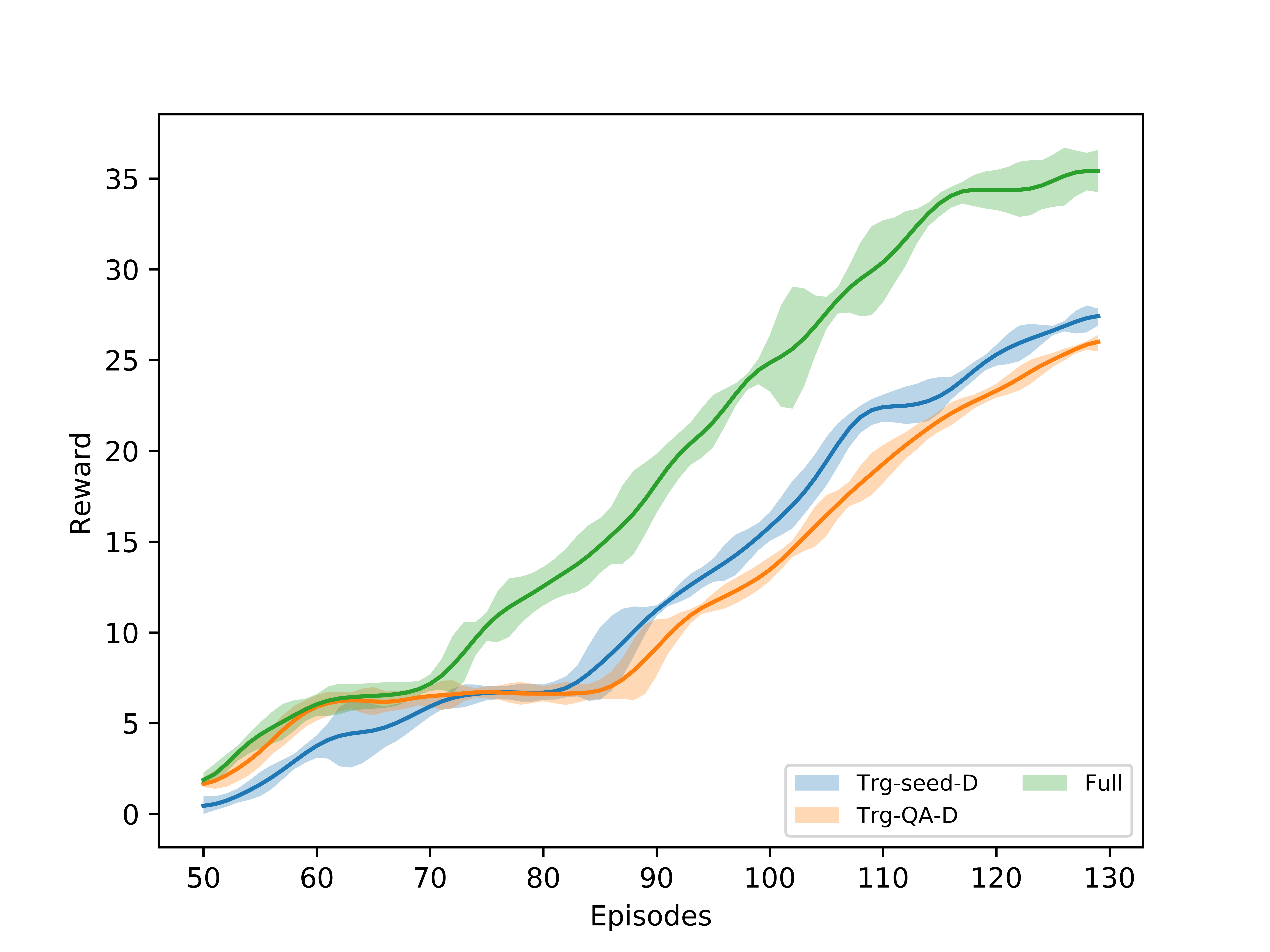}
    \caption{Reward curve for select experiments in the horror domain.}
    \label{fig:reward-horror}
\end{figure}

\begin{table*}[]
\centering
\footnotesize
\caption{Results for the slice of life games. ``KG-DQN Full'' refers to KG-DQN when seeded first, trained on the source, then transferred to the target. All experiment with QA indicate pre-training. S, D indicate sparse and dense reward respectively.}
\label{table:results-sol}
\begin{tabular}{c|c|c|c}
Experiment                  & Init. Rwd. & Final Rwd. & Steps \\ \hline
\multicolumn{4}{l}{Source Game ({\em TextWorld})}\\
\hline
KG-DQN no transfer                       &      2.6 $\pm$ 0.73        &      4.7 $\pm$ 0.23        &   110.83 $\pm$ 4.92    \\
KG-DQN w/ QA                  &       2.8 $\pm$ 0.61       &      4.9 $\pm$ 0.09        &    88.57 $\pm$ 3.45   \\
KG-DQN seeded                 &      3.2 $\pm$ 0.57        &      4.8 $\pm$ 0.16       &    91.43 $\pm$ 1.89   \\
\hline
\multicolumn{4}{l}{Target Game ({\em 9:05})}\\
\hline
KG-DQN untuned (D)                      &      -       &       2.5 $\pm$0.48      &    1479.0  $\pm$ 22.3 \\
KG-DQN no transfer (S)                      &      -       &       -      &    1916.0  $\pm$ 33.17 \\
KG-DQN no transfer (D)                      &    0.8   $\pm$ 0.32      &      16.5  $\pm$ 1.58     &  1267.2 $\pm$ 7.5  \\
KG-DQN w/ QA (S)                  &       -      &       -      &  1428.0 $\pm$  11.26  \\
KG-DQN w/ QA (D)                  &   1.3   $\pm$ 0.24     &     17.4   $\pm$ 1.84     &  1127.0 $\pm$  31.22  \\
KG-DQN seeded (D)       &     1.4 $\pm$ 0.35         &    16.7 $\pm$ 2.41        &  1393.33  $\pm$ 26.5     \\
KG-DQN Full (D)       &     2.7  $\pm$ 0.65      &    19.7  $\pm$    2.0    &  274.76  $\pm$  21.45\\
\hline
\end{tabular}
\end{table*}

\begin{table*}[]
\footnotesize
\centering
\caption{Results for horror games. Note that the reward type is dense for all results. ``KG-DQN Full`` refers to KG-DQN seeded, transferred from source. All experiment with QA indicate pre-training.}
\label{table:results-horror}
\begin{tabular}{c|c|c|c}
Experiment                  & Init. Rwd. & Final Rwd. & Steps \\ \hline
\multicolumn{4}{l}{Source Game ({\em Afflicted})}\\
\hline
KG-DQN no transfer                &     3.0 $\pm$ 1.3         &      14.1 $\pm$ 1.73        &   1934.7 $\pm$ 85.67    \\
KG-DQN w/ QA         &      4.3 $\pm$ 1.34        &     15.1 $\pm$ 1.60         &  1179 $\pm$ 32.07     \\
KG-DQN seeded          &       4.1 $\pm$ 1.19       &  14.6 $\pm$ 1.26            &   1125.3 $\pm$ 49.57    \\
\hline
\multicolumn{4}{l}{Target Game ({\em Anchorhead})}\\
\hline
KG-DQN untuned                 &      -        &    3.8 $\pm$ 0.23         &   -    \\
KG-DQN no transfer                 &      1.0 $\pm$ 0.34        &    6.8 $\pm$ 0.42         &   -    \\
KG-DQN w/ QA           &       3.6 $\pm$ 0.91       &      24.8 $\pm$ 0.6        &   4874 $\pm$ 90.74    \\
KG-DQN seeded          &      1.7 $\pm$ 0.62        &    26.6 $\pm$ 0.42          &  4937 $\pm$   42.93 \\
KG-DQN full &      4.1 $\pm$ 0.9         &    39.9 $\pm$ 0.53         &      4334.3 $\pm$ 56.13\\
\hline
\end{tabular}
\end{table*}

The structure of the experiments are such that the for each of the domains, the target task game is more complex that the source task game.
The slice of life games are also generally less complex than the horror games; they have a simpler vocabulary and a more linear quest structure.
Additionally, given the nature of interactive fiction games, it is nearly impossible---even for human players---to achieve completion in the minimum number of steps (as given by the steps to completion in Table~\ref{table:games});
each of these games are puzzle based and require extensive exploration and interaction with various objects in the environment to complete.

Table~\ref{table:results-sol} and Table~\ref{table:results-horror} show results for the slice of life and horror domains, respectively.
In both domains seeding and QA pre-training improve performance by similar amounts from the baseline on both the source and target task games. 
A series of t-tests comparing the results of the pre-training and graph seeding with the baseline KG-DQN show that all results are significant with $p<0.05$.
Both the pre-training and graph seeding perform similar functions in enabling the agent to explore more effectively while picking high utility actions.

Even when untuned, i.e. evaluating the agent on the target task after having only trained on the source task, the agent shows better performance than training on the target task from scratch using the sparse reward.
As expected, we see a further gain in performance when the dense reward function is used for both of these domains as well.
In the horror domain, the agent fails to converge to a state where it is capable of finishing the game without the dense reward function due to the horror games being more complex.

When an agent is trained using on just the target task horror game, {\em Anchorhead}, it does not converge to completion and only gets as far as achieving a reward of approximately 7 (max. observed reward from the best model is 41).
This corresponds to a point in the game where the player is required to use a term in an action that the player has never observed before, ``look up Verlac'' when in front of a certain file cabinet---``Verlac`` being the unknown entity.
Without seeding or QA pre-training, the agent is unable to cut down the action space enough to effectively explore and find the solution to progress further.
The relative effectiveness of the gains in initial reward due to seeding appears to depend on the game and the corresponding static text document. 
In all situations except Anchohead, seeding provides comparable gains in initial reward as compared to QA --- there is no statistical difference between the two when performing similar t-tests.

When the full system is used---i.e. we seed the knowledge graph, pre-train QA, then train the source task game, then the target task game using the augmented reward function---we see a significant gain in performance, up to an 80\% gain in terms of completion steps in some cases.
The bottleneck at reward 7 is still difficult to pass, however, as seen in Fig.~\ref{fig:reward-horror}, in which we can see that the agent spends a relatively long time around this reward level unless the full transfer technique is used.
We further see in Figures~\ref{fig:reward-sol},~\ref{fig:reward-horror} that transferring knowledge results in the agent learning this higher quality policy much faster.
In fact, we note that training a full system is more efficient than just training the agent on a single task, i.e. training a QA system then a source task game for 50 episodes then transferring and training a seeded target task game for 50 episodes is more effective than just training the target task game by itself for even 150+ episodes.

\section{Conclusions}

We have demonstrated that using knowledge graphs as a state representation enables efficient transfer between deep reinforcement learning agents designed to play text-adventure games, reducing training times and increasing the quality of the learned control policy.
Our results show that we are able to extract a graph from a general static text resource and use that to give the agent knowledge regarding domain specific vocabulary, object affordances, etc.
Additionally, we demonstrate that we can effectively transfer knowledge using deep Q-network parameter weights, either by pre-training portions of the network using a question-answering system or by transferring parameters from a source to a target game.
Our agent trains faster overall, including the number of episodes required to pre-train and train on a source task, and performs up to 80\% better on convergence than an agent not utilizing these techniques.

We conclude that knowledge graphs enable transfer in deep reinforcement learning agents by providing the agent with a more explicit--and interpretable--mapping between the state and action spaces of different games.
This mapping helps overcome the challenges twin challenges of partial observability and combinatorially large action spaces inherent in all text-adventure games by allowing the agent to better explore the state-action space.

\section{Acknowledgements}
This  material  is  based  upon  work  supported  by  the  National  Science  Foundation  under  Grant  No.  IIS-1350339. Any  opinions,  findings,  and  conclusions  or  recommendations expressed in this material are those of the author(s) and do not necessarily reflect the views of the National Science Foundation.

\bibliography{emnlp-ijcnlp-2019.bib}
\bibliographystyle{acl_natbib}

\end{document}